# Variational Recurrent Neural Machine Translation

**Jinsong Su**[1], **Shan Wu**[1,2], **Deyi Xiong**[3*], **Yaojie Lu**[2], **Xianpei Han**[2], **Biao Zhang**[1]

Xiamen University, Xiamen, China[1]
Institute of Software, Chinese Academy of Sciences, Beijing, China[2]
Soochow University, Suzhou, China[3]
jssu@xmu.edu.cn, wushan@stu.xmu.edu.cn, dyxiong@suda.edu.cn
yaojie2017@iscas.ac.cn, xianpei@nfs.iscas.ac.cn, zb@stu.xmu.edu.cn

### Abstract

Partially inspired by successful applications of variational recurrent neural networks, we propose a novel variational recurrent neural machine translation (VRNMT) model in this paper. Different from the variational NMT, VRNMT introduces a series of latent random variables to model the translation procedure of a sentence in a generative way, instead of a single latent variable. Specifically, the latent random variables are included into the hidden states of the NMT decoder with elements from the variational autoencoder. In this way, these variables are recurrently generated, which enables them to further capture strong and complex dependencies among the output translations at different timesteps. In order to deal with the challenges in performing efficient posterior inference and large-scale training during the incorporation of latent variables, we build a neural posterior approximator, and equip it with a reparameterization technique to estimate the variational lower bound. Experiments on Chinese-English and English-German translation tasks demonstrate that the proposed model achieves significant improvements over both the conventional and variational NMT models.

## 1. Introduction

Recently, neural machine translation (NMT) has gradually established state-of-the-art results over statistical machine translation (SMT) on various language pairs. Most NMT models consist of two recurrent neural networks (RNNs): a bidirectional RNN based encoder that transforms source sentence $\mathbf{x} = \{x_1, x_2...x_{T_x}\}$ into a hidden state sequence, and a decoder that generates the corresponding target sentence $\mathbf{y} = \{y_1, y_2...y_{T_y}\}$ by exploiting source-side contexts via an attention network (Bahdanau, Cho, and Bengio 2015). This attentional neural encoder-decoder framework has now become the dominant architecture for NMT.

Within this framework, semantic representations of source and target sentences are learned in an implicit way. As a result, the learned semantic representations are far from being sufficient for capturing all semantic details and dependencies (Sutskever, Vinyals, and Le 2014; Tu et al. 2016). To complement the insufficiency of semantic representations of NMT, Zhang et al. (2016) present variational NMT (VNMT) which incorporates a latent random variable into NMT, serving as a global semantic signal for generating good translations. However, the internal transition structure of RNN is entirely deterministic, and hence, this implementation may not be an effective way to model high variability observed in structured data, such as language modeling and machine translation (Chung et al. 2015). Therefore, the potential of VNMT is limited and how to better improve NMT with latent variables is still open for further exploration.

In this paper, we propose a variational recurrent NMT (VRNMT) model to deal with the above-mentioned problem, motivated by recent success of the variational recurrent neural network (VRNN) (Chung et al. 2015). It is illustrated in Fig. 1. VRNMT explicitly models underlying semantics of bilingual sentence pairs, which are then exploited to refine translation. However, instead of only employing a single latent variable to capture the global semantics of each parallel sentence, we assume that there is a continuous latent random variable sequence $\mathbf{z} = \{z_1, z_2..., z_{T_y}\}$ in the underlying semantic space, where the iteratively generated variable $z_j$ participates in the generations of each target word $y_j$ and hidden state $s_{j+1}$. Formally, the conditional probability $p(\mathbf{y}|\mathbf{x})$ is decomposed as follows:

$$p(\mathbf{y}|\mathbf{x}) = \prod_{j=1}^{T_y} p(y_j|\mathbf{x}, y_{<j}) = \prod_{j=1}^{T_y} \int_{z_j} p(y_j, z_j|\mathbf{x}, y_{<j}) dz_j$$

$$= \prod_{j=1}^{T_y} \int_{z_j} p(y_j|\mathbf{x}, y_{<j}, z_j) p(z_j|\mathbf{x}, y_{<j}) dz_j \quad (1)$$

where $z_j$ encodes the semantic contexts at the $j$-th timestep. In doing so, we expect these latent variables to efficiently model the strong and complex dependencies between adjacent target words, which may not be effectively and sufficiently captured by the conventional NMT or VNMT.

However, the incorporation of latent variables into the existing NMT models faces two challenges, as mentioned in (Zhang et al. 2016): 1) the posterior inference in our model is intractable; 2) large-scale training, which lays the ground for the data-driven NMT, is accordingly problematic. To address these two issues, we follow Zhang et al. (2016) to use deep neural networks, which are capable of learning highly nonlinear functions, to fit the latent-variable-related distributions, i.e. the prior and posterior. The former is $p_\theta(z_j|\mathbf{x}, y_{<j})$, conditioned on source sentence and previously generated target words. The latter is $q_\phi(z_j|\mathbf{x}, y_{\leq j})$, approximated from all observed variables. To enable efficient inference and learning, we adopt the *reparameterization trick* (Kingma and Welling 2014; Rezende, Mohamed, and Wierstra 2014) to bridge the gap between these two distributions. In this way, our model becomes an end-to-end neural network endowed with the stochastic optimization ability for enhancing its generality.

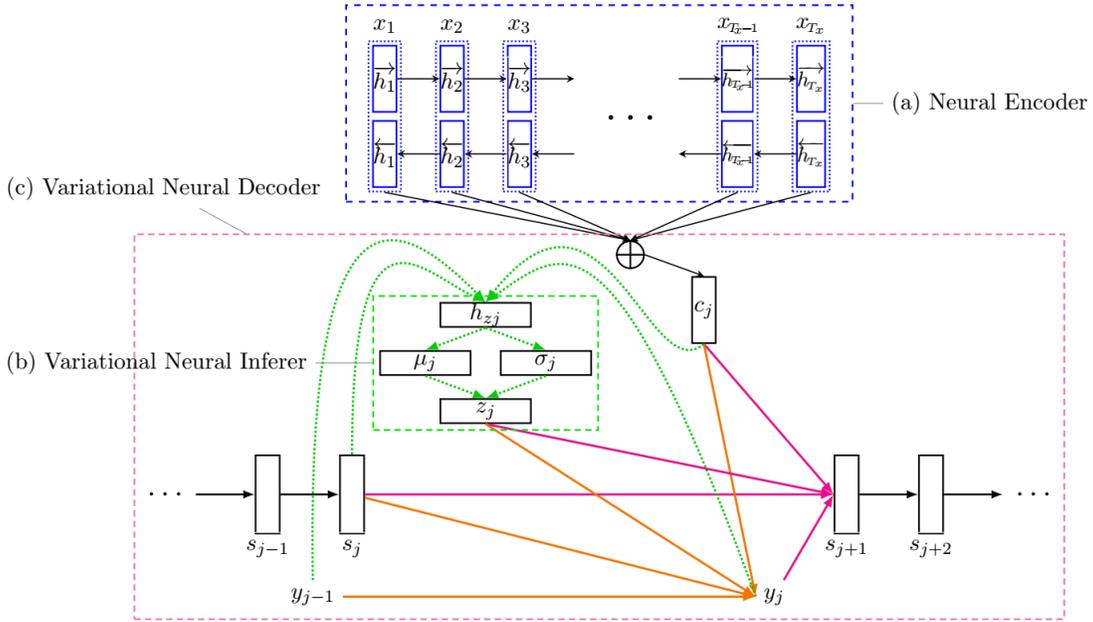

Figure 1: Illustration of VRNMT as a directed graph. The green dotted lines illustrate the modeling procedure of the latent variable $z_j$. The orange lines show the information flow for the prediction of target word $y_j$. The red lines highlight the incorporation of $z_j$ into the encoding of hidden state vector $s_{j+1}$.

Our main contributions in this work are twofold:

- We propose a VRNMT model that not only explores the utilization of high-level latent random variables but also efficiently captures the strong and complex dependencies between neighboring target words for NMT. To the best of our knowledge, this is the first attempt to adapt VRNN into NMT modeling.
- Experimental results on Chinese-English and English-German translation tasks show that the proposed model significantly outperforms the conventional NMT and VNMT models.

## 2. Background

In this section, we briefly describe the attention-based NMT model and VRNN, which provide background knowledge for the proposed model.

### 2.1 Attention-based NMT Model

Currently, the dominant NMT model mainly consists of a neural encoder and a neural decoder with an attention network (Bahdanau, Cho, and Bengio 2015).

Generally, the encoder is a bidirectional RNN learning hidden representations of a source sentence in the forward and backward directions. The learned hidden states in two directions are then concatenated to form source annotations $\{h_i = [\overrightarrow{h}_i^T, \overleftarrow{h}_i^T]^T\}$, where $h_i$ encodes the contextual semantics of the $i$-th word with respect to all other surrounding source words.

Likewise, the decoder is a forward RNN that adopts the nonlinear function $g(\cdot)$ to sequentially generate the translation $\mathbf{y}$ as $p(y_j|\mathbf{x}, y_{<j})=g(y_{j-1}, s_j, c_j)$, where $s_j$ and $c_j$ denote the decoding state and the source context at the $j$-th timestep, respectively. Among them, $s_i$ is computed as $s_j=GRU(s_{j-1}, y_{j-1}, c_j)$. Here we use GRU for both the encoder and decoder in this work. However, our work is also applicable to other types of RNNs. According to the attention mechanism, we calculate $c_j$ as the weighted sum of the source annotations $\{h_i\}$:

$$c_j = \sum_{i=1}^{T_x} \alpha_{j,i} \cdot h_i, \quad (2)$$

where $\alpha_{j,i}$ evaluates how well $y_j$ and $h_i$ match, computed as follows:

$$\alpha_{j,i} = \frac{exp(e_{j,i})}{\sum_{i'=1}^{T_x} exp(e_{j,i'})}, \quad (3)$$

$$e_{j,i} = v_a^T tanh(W_a s_{j-1} + U_a h_i), \quad (4)$$

where $W_a, U_a$ and $v_a$ are the weight matrices of the attention model.

### 2.2 VRNN

VRNN is a recurrent extension of the conventional VAE (Chung et al. 2015). As shown in Fig. 2, it contains a VAE at every timestep and introduces four kinds of operations to explicitly model the dependencies between latent random variables across subsequent timesteps.

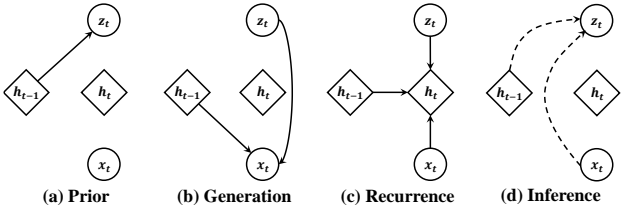

Figure 2: Graphical illustrations of each operation of the VRNN.

**Prior, Generation and Recurrence**. Based on the hidden state $h_{t-1}$ of the RNN, VRNN first produces a latent semantic variable $z_t$, which is then used to guide the generation of the hidden state $h_t$ and word $x_t$ at the $t$-th timestep. In doing so, the temporal structure of sequential data is exploited for VRNN modeling.

Different from the standard VAE where the prior on the latent random variable follows a standard Gaussian distribution, VRNN assumes that $z_t$ obeys the following Gaussian with the parameters $\mu_{0,t}$ and $\sigma_{0,t}$

$$z_t \sim \mathcal{N}(\mu_{0,t}, \text{diag}(\sigma_{0,t}^2)), \quad (5)$$

where $\mu_{0,t}$ and $\sigma_{0,t}$ can be produced by any highly flexible neural networks. Moreover, the generation distribution of $x_t$ will be conditioned on both $z_t$ and $h_{t-1}$ such that

$$x_t|z_t \sim \mathcal{N}(\mu_{x,t}, \text{diag}(\sigma_{x,t}^2)), \quad (6)$$

where $\mu_{x,t}$ and $\sigma_{x,t}$ are the parameters of the generation distribution. Note that they can also be computed by any highly flexible neural network.

Then, we introduce $z_t$ to update the hidden state $h_t$ in a recurrent way

$$h_t = f_\theta(h_{t-1}, x_t, z_t). \quad (7)$$

Finally, the parameterization of the generative model can be factorized as follows:

$$p_\theta(\mathbf{x}) = \int_\mathbf{z} p_\theta(\mathbf{x}, \mathbf{z}) d\mathbf{z}, \quad (8)$$

$$p_\theta(\mathbf{x}, \mathbf{z}) = \prod_{t=1}^T p(x_t|x_{<t}, z_{\leq t}) p(z_t|x_{<t}, z_{<t}), \quad (9)$$

where $T$ is the sequence length.

**Inference**. Similarly, the approximate posterior is defined as a function of both $x_t$ and $h_t$

$$z_t|x_t \sim \mathcal{N}(\mu_{z,t}, \text{diag}(\sigma_{z,t}^2)), \quad (10)$$

where $\mu_{z,t}$ and $\sigma_{z,t}$ are the parameters of the approximate posterior. In doing so, $h_t$ enables the encoding of the approximate posterior and the decoding for generation are closely tied.

Finally, the objective function becomes a timestep-wise variational lower bound

$$\mathcal{L}_{\text{VRNN}}(\theta, \phi; \mathbf{x}) = \mathbb{E}_{q(z_{\leq M}|x_{\leq T})}[\sum_{t=1}^T (-KL(q(z_t|x_{\leq t}, z_{<t})$$
$$||p(z_t|x_{<t}, z_{<t})) + \log p(x_t|z_{\leq t}, x_{<t}))] \quad (11)$$

As implemented in VAE, we maximize the variational lower bound with respect to their parameters to jointly learn the generative and inference models. Note that we also have to model the posterior $p_\theta(z_t|x_{\leq t}, z_{<t})$ while the integration of $z_t$ still leads to difficulties in the posterior inference and large-scale learning. Likewise, we adopt the neural approximation and reparameterization trick to handle this issue.

## 3. Our Model

In this section, we extend VNMT into VRNMT by adapting VRNN into NMT. In VRNMT, the semantic dependencies between adjacent target words can be captured to refine translation. Formally, the variational lower bound of VRNMT is defined as follows:

$$\mathcal{L}_{\text{VRNMT}}(\mathbf{y}|\mathbf{x}; \theta, \phi) = \sum_{j=1}^{T_y} \mathcal{L}_{\text{VRNMT}}(y_j|\mathbf{x}, y_{<j}; \theta, \phi)$$

$$= \sum_{j=1}^{T_y} \{ -KL(q_\phi(z_j|\mathbf{x}, y_{\leq j})||p_\theta(z_j|\mathbf{x}, y_{<j}))$$
$$+ \mathbb{E}_{q_\phi(z_j|\mathbf{x}, y_{\leq j})}[\log p_\theta(y_j|\mathbf{x}, y_{<j}, z_j)] \}, \quad (12)$$

where $p_\theta(z_j|\mathbf{x}, y_{<j})$ is the prior, $q_\phi(z_j|\mathbf{x}, y_{\leq j})$ is the approximated posterior, and $p_\theta(y_j|\mathbf{x}, y_{<j}, z_j)$ is the generation model.

As shown in Eq. (12), VRNMT mainly contains three neural network based components: (1) a neural encoder for encoding source sentences, (2) a variational neural inferer for $q_\phi(z_j|\mathbf{x}, y_{\leq j})$ and $p_\theta(z_j|\mathbf{x}, y_{<j})$, and (3) a variational neural decoder that models $p_\theta(y_j|\mathbf{x}, y_{<j}, z_j)$.

### 3.1 Neural Encoder

The encoder of VRNMT is the same as that of the conventional NMT. Due to the limitation of space, we omit the description of the VRNMT encoder (See Section 2.1 for reference).

### 3.2 Variational Neural Inferer

As described previously, the key of variational models lies in how to model the distributions related to latent random variables. With respect to VRNMT, we focus on how to model the posterior $q_\phi(z_j|\mathbf{x}, y_{\leq j})$ and the prior $p_\theta(z_j|\mathbf{x}, y_{<j})$.

**The Posterior Model.** Under the assumption that the posterior $q_\phi(z_j|\mathbf{x}, y_{\leq j})$ follows the multivariate Gaussian distribution with a diagonal covariance structure, we apply neural networks to simulate the posterior model. Concretely, we compute $q_\phi(z_j|\mathbf{x}, y_{\leq j})$ as

$$q_\phi(z_j|\mathbf{x}, y_{\leq j}) = \mathcal{N}(z_j; \mu_j(\mathbf{x}, y_{\leq j}), \sigma_j(\mathbf{x}, y_{\leq j})^2 \mathbf{I}). \quad (13)$$

As illustrated in Fig. 1, the mean $\mu_j$ and standard derivation $\sigma_j$ of neural networks are imposed on $\mathbf{x}$ and $y_{\leq j}$.

Obviously, the key to estimate $z_j$ is how to calculate $\mu_j$ and $\sigma_j$. To this end, we first apply the element-wise activation function $g(\cdot)$ to perform a nonlinear transformation projecting $y_{j-1}$, $s_j$, $c_j$ and $y_j$ onto our latent semantic space:

$$\mathbf{h}_{zj} = g(W_z^\phi [y_{j-1}; s_j; c_j; y_j] + b_z^\phi). \quad (14)$$

where $W_z^\phi$ and $b_z^\phi$ are the parameter matrix and bias term, respectively. Finally, we introduce linear regressions with parameters $W_\mu^\phi, W_\sigma^\phi, b_\mu^\phi$ and $b_\sigma^\phi$ to obtain $d_z$-dimension vectors $\mu_j$ and $\log\sigma_j^2$ as follows:

$$\mu_j = W_\mu^\phi \mathbf{h}_{zj} + b_\mu^\phi \quad (15)$$

$$\log\sigma_j^2 = W_\sigma^\phi \mathbf{h}_{zj} + b_\sigma^\phi \quad (16)$$

To obtain a representation for latent variable $z_j$, we follow the implementation of VAE to reparameterize it as $z_j = \mu_j + \sigma_j \odot \epsilon, \epsilon \sim \mathcal{N}(0, \mathbf{I})$. Intuitively, this reparameterization procedure bridges the gap between $p_\theta(y_j|\mathbf{x}, y_{<j}, z_j)$ and $q_\phi(z_j|\mathbf{x}, y_{\leq j})$. Thus, our model is an end-to-end neural network endowed with the generality ability.

**The Prior Model.** Except for the absence of $y_j$, the neural model for the prior $p_\theta(z_j|\mathbf{x}, y_{<j})$ is identical to that (i.e. Eq (13)) for the posterior $q_\phi(z_j|\mathbf{x}, y_{\leq j})$. Here we still model the prior as a multivariate Gaussian distribution but introduce different parameters for the prior and the posterior:

$$p_\theta(z_j|\mathbf{x}, y_{<j}) = \mathcal{N}(z_j; \mu_j'(\mathbf{x}, y_{<j}), \sigma_j'(\mathbf{x}, y_{<j})^2 \mathbf{I}), \quad (17)$$

Using the similar way in computing the posterior model, we first calculate $\mathbf{h}'_{zj}$ without $y_j$ in the following way:

$$\mathbf{h}'_{zj} = g(W_z^\theta [y_{j-1}; s_j; c_j] + b_z^\theta). \quad (18)$$

Here $W_z^\theta$ and $b_z^\theta$ are the parameter matrix and bias term, respectively.

Then, we use $\mathbf{h}'_{zj}$ to generate the mean $\mu'_j$ and standard derivation $\log\sigma_j^{'2}$ with the parameters $W_*^\theta$ and $b_*^\theta$:

$$\mu'_j = W_\mu^\theta \mathbf{h}'_{zj} + b_\mu^\theta \quad (19)$$

$$\log\sigma_j^{'2} = W_\sigma^\theta \mathbf{h}'_{zj} + b_\sigma^\theta \quad (20)$$

Different from the posterior model, we directly set $z_j$ as $\mu'_j$, as implemented in (Zhang et al. 2016). Note that we also introduce noises to generate non-fixed representation $z_j$ in practice, which enables our model to avoid overfitting to some extent.

Finally, $z_j$ is integrated into our decoder to improve translation. The details will be illustrated in the following subsection.

### 3.3 Variational Neural Decoder

Given the source sentence $\mathbf{x}$, the previously generated target words $y_{<j}$, and the semantic latent variable $z_j$, we compute the probability distribution over the translation $y_j$ as

$$p(y_j|\mathbf{x}, y_{<j}, z_j) = g_{\text{VRNMT}}(y_{j-1}, s_j, c_j)$$
$$\propto \exp\{g(W_d[y_{j-1}; s_j; c_j; z_j] + b_d)\} \quad (21)$$

Unlike the conventional NMT, we first produce $z_j$ using $y_{j-1}, s_j$, and $c_j$ (see the dashed green lines in Fig. 1), and then integrate $z_j$ with $y_{j-1}, s_j$, and $c_j$ to generate translation probability distribution (see the orange lines in Fig. 1). Besides, we use $z_j$ to generate the next hidden state (see the red lines in Fig. 1). Formally, the GRU transition equations of our decoder are as follows:

$$r_{j+1} = \sigma(W_r \vec{y}_j + U_r s_j + C_r c_j + V_r z_j + b_r) \quad (22)$$

$$u_{j+1} = \sigma(W_u \vec{y}_j + U_u s_j + C_u c_j + V_u z_j + b_u) \quad (23)$$

$$\tilde{s}_{j+1} = \tanh(W \vec{y}_j + U[r_{j+1} \odot s_j] + Cc_j + Vz_j + b) \quad (24)$$

$$s_{j+1} = (1 - u_{j+1}) \odot s_j + u_{j+1} \odot \tilde{s}_{j+1}, \quad (25)$$

where $W_*, U_*, C_*, V_*$, and $b_*$, are the model parameters of GRU in VRNMT. Particularly, we initialize the hidden state $s_0$ in a way similar to (Bahdanau, Cho, and Bengio 2015).

It should be noted that the latent semantic variable $z_j$ has an important influence on the representation of hidden state $s_{j+1}$ through the gates $r_{j+1}$ and $u_{j+1}$, and temporary hidden state $\tilde{s}_{j+1}$. This allows our model to access the semantic information of $z_j$ indirectly since the prediction of $y_{j+1}$ depends on $s_{j+1}$. On the other hand, $s_{j+1}$, in turn will constrain the generation of $z_{j+1}$ at the next timestep. Therefore, the context dependencies between adjacent timesteps are indirectly exploited to refine translation.

### Model Training

The final objective for one bilingual sentence $(\mathbf{x}, \mathbf{y})$ involves the following two parts: $-\text{KL}(q_\phi(z_j|\mathbf{x}, y_{\leq j})||p_\theta(z_j|\mathbf{x}, y_{<j}))$ and $\mathbb{E}_{q_\phi(z_j|\mathbf{x}, y_{\leq j})}[\cdot]$. We also apply the Monte Carlo method to approximate $\mathbb{E}_{q_\phi(z_j|\mathbf{x}, y_{\leq j})}[\cdot]$. Formally, the joint training objective becomes

$$\mathcal{L}_{RLV}(\theta, \phi; \mathbf{x}, \mathbf{y}) = \sum_{j=1}^{T_y} \{ -\text{KL}(q_\phi(z_j|\mathbf{x}, y_{\leq j})||p_\theta(z_j|\mathbf{x}, y_{<j}))$$
$$+ \mathbb{E}_{q_\phi(z_j|\mathbf{x}, y_{\leq j})}[\log p_\theta(y_j|\mathbf{x}, y_{<j}, z_j)] \}$$
$$\simeq \sum_{j=1}^{T_y} \{ -\text{KL}(q_\phi(z_j|\mathbf{x}, y_{\leq j})||p_\theta(z_j|\mathbf{x}, y_{<j}))$$
$$+ \frac{1}{L} \sum_{l=1}^{L} \log p_\theta(y_j|\mathbf{x}, y_{<j}, z_j^{(l)}) \} \quad (26)$$

where $z_j^{(l)} = \mu_j + \sigma_j \odot \epsilon^{(l)}, \epsilon^{(l)} \sim \mathcal{N}(0, \mathbf{I})$, and $L$ is the number of samples. Essentially, VRNMT can be considered as a regularized version of NMT, which introduces noise $\epsilon^{(l)}$ at each timestep to enhance its robustness. Notice that both the KL divergence and the approximate expectation are differentiable. Therefore, we can jointly optimize the model parameters $\theta$ and variational parameters $\phi$ using standard gradient ascent.

## 4. Experiments

We conducted experiments on Chinese-English and English-German translation to examine the effectiveness of our model.

### 4.1 Setup

Our Chinese-English training data consists of 1.25M LDC sentence pairs, with 27.9M Chinese words and 34.5M English words respectively. We used the NIST MT02 dataset as the validation set, and the NIST MT03/04/05/06 datasets as the test sets. In English-German translation, our training data consists of 4.46M sentence pairs with 116.1M English words and 108.9M German words. We used the news-test 2013 as the validation set and the news-test 2015 as the test set. Following Sennrich et al. (2016), we adopted byte pair encoding to segment words into subwords for English-German translation. Finally, we used BLEU (Papineni et al. 2002) as our evaluation metric, and performed paired bootstrap sampling (Koehn 2004) for statistical significance test using the Moses script.

We set the maximum length of training sentences to be 50 words, and preserved the most frequent 30K (Chinese-English) and 50K (English-German) words as both the source and target vocabulary, covering approximately 97.4%/100.0% and 99.3%/98.2% on the source/target side of the two parallel corpora respectively. All other words were replaced with a specific token "UNK". We applied *Rmsprop* (Graves 2013) with iterNum=5, momentum=0, $\rho$=0.95, and $\epsilon$=1×10$^{-4}$ to train various NMT models. The settings of our model were the same as in (Bahdanau, Cho, and Bengio 2015), except for some hyper-parameters specific to our model. Specifically, we set word embedding dimension as 620, hidden layer size as 1000, learning rate as 5×10$^{-4}$, batch size as 80, gradient norm as 1.0, and dropout rate as 0.3. Particularly, we initialized the parameters of VRNMT with the trained conventional NMT model. As implemented in VAE, we set the sampling number $L$=1, and $d'_e$=$d_z$=2$d_f$=2000 according to preliminary experiments. During decoding, we used the beam-search algorithm, and set beam sizes of all models as 10.

### 4.2 Systems for Comparison

We compared our model against the following systems:

(1) **Moses**[1]. An open source phrase-based SMT system with default settings and a 4-gram language model trained on the target portion of the training data.

(2) **DL4MT**. Our re-implementation of the attention-based NMT system (Bahdanau, Cho, and Bengio 2015) with slight changes from dl4mt tutorial[2].

(3) **VNMT**. It is a variational NMT system (Zhang et al. 2016) that incorporates a continuous latent variable to model the underlying semantics of sentence pairs.

(4) **VRNMT(-TD)**. A variant of our model without introducing temporal dependencies between the latent random variables. It differs from our model in that the input of posterior model contains only $y_j$ but not $y_{j-1}, s_j, c_j$. More specifically, we removed $y_{j-1}, s_j$, and $c_j$ from Eq. (14). Thus, the latent variables of VRNMT(-TD) directly obey the standard Gauss distribution rather than depend on the output at the previous timestep. As we incorporate temporal dependencies into the prior, we will directly study the impact of the latent random variables on modeling variability characterized by dependencies among output words in comparison to VRNMT(-TD).

---
[1]http://www.statmt.org/moses/
[2]https://github.com/nyu-dl/dl4mt-tutorial

| System | MT03 | MT04 | MT05 | MT06 | Ave. |
|---|---|---|---|---|---|
| COVERAGE | 34.49 | 38.34 | 34.91 | 34.25 | 35.50 |
| InterAtten | 35.09 | 37.73 | 35.53 | 34.32 | 35.67 |
| MemDec | 36.16 | 39.81 | 35.91 | 35.98 | 36.97 |
| DMAtten | **38.33** | 40.11 | 36.71 | 35.29 | 37.61 |
| Moses | 32.93 | 34.76 | 31.31 | 31.05 | 32.51 |
| DL4MT | 36.59 | 39.57 | 35.56 | 35.29 | 36.75 |
| VNMT | 37.23 | 40.32 | 36.28 | 35.73 | 37.39 |
| VRNMT(-TD) | 36.97 | 40.07 | 36.13 | 35.49 | 37.17 |
| VRNMT | 38.08$^{*}_{++}$ | 41.07$^{**}_{++}$ | 36.82$^{**}_{++}$ | 36.72$^{*}_{++}$ | 38.17 |

Table 1: Case-insensitive BLEU scores of Chinese-English translation. $*/**$ and $+/++$: significant over VNMT and VRNMT(-TD) at 0.05/0.01, respectively. **COVERAGE** (Tu et al. 2016) presented a coverage model to alleviate the over-translation and under-translation problems. **InterAtten** (Meng et al. 2016) exploited a readable and writable attention mechanism to record interactive history in decoding. **MemDec** (Wang et al. 2016) introduced external memory to improve translation quality. **DMAtten** (Zhang et al. 2017) explicitly incorporated the word reordering knowledge into the attention model of NMT. Note that all these studies focus on capturing semantic information for NMT.

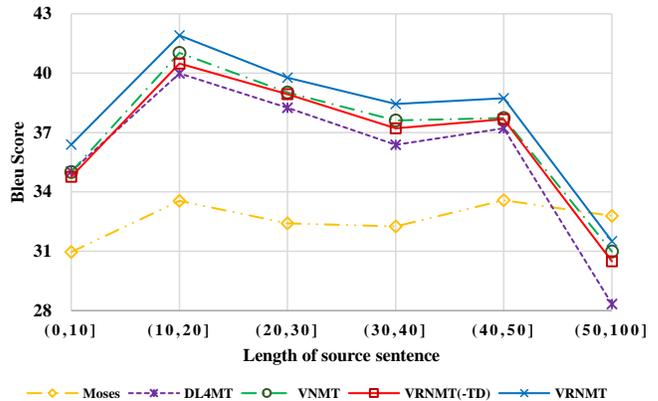

Figure 3: BLEU scores over different lengths of translated sentences.

### 4.3 Results on Chinese-English Translation

In addition to the above systems for comparison, we also displayed the BLEU scores of several recent NMT models (Tu et al. 2016; Meng et al. 2016; Wang et al. 2016; Zhang et al. 2017) that have been trained on the same training corpus as ours.

Table 1 shows case-insensitive BLEU scores on Chinese-English datasets. Overall, VRNMT significantly improves translation quality on all test sets, achieving the gains of 5.66, 1.42, 0.78 and 1.0 BLEU points over Moses, DL4MT, VNMT and VRNMT(-TD), respectively. Compared to the existing NMT models, VRNMT is better than them as shown in Table 1. These results echo the results reported in (Zhang et al. 2016), indicating the integration of latent variables is effective for improving NMT. Particularly, VRNMT performs significantly better than VRNMT(-TD), indicating that explicitly modeling the temporal dependencies between latent random variables indeed further benefits NMT.

| Source | píngrǎng cǎiqǔ shàngshù xíngdòng zhīhòu sì tiān , liánhéguó ānquán lǐshìhuì de wǔ gè chángrèn lǐshìguó dōu wèi cǐ yī wēijī cǎiqǔ yùfángxìng wàijiāo xíngdòng . |
|---|---|
| Reference | *four days after pyongyang adopted the aforesaid action , the five permanent members of united nations security council have all taken preemptive diplomatic actions for the crisis .* |
| Moses | *pyongyang by four days after the operation of the un security council , the five permanent members to adopt preventive diplomacy this crisis .* |
| DL4MT | *the four permanent member states of the united nations security council and the five permanent members of the un security council have adopted a preventive diplomatic action following the four - day operation .* |
| VNMT | *four days after north korea took the above actions , the five permanent members of the un security council have adopted preventive diplomatic activities .* |
| VRNMT(-TD) | *four permanent members of the security council of the united nations security council have taken preventive diplomatic actions during the four - day period following the above actions .* |
| VRNMT | *four days after pyongyang took the action , the five permanent members of the un security council have adopted preventive diplomatic actions for the crisis .* |

Table 2: Translation examples of different systems. Words highlighted in underlines are not fluently translated, in wavy lines are incorrectly translated, in dashed lines are over-translated, and in dotted lines are under-translated.

| System | 1-Gram | 2-Gram | 3-Gram | 4-Gram |
|---|---|---|---|---|
| Reference | 12.94 | 1.80 | 0.93 | 1.29 |
| DL4MT | 19.62 | 5.34 | 2.96 | 2.31 |
| VNMT | 19.45 | 5.24 | 2.93 | 2.29 |
| VRNMT(-TD) | 19.54 | 5.25 | 2.93 | 2.35 |
| VRNMT | **18.83** | **4.97** | **2.90** | **2.25** |

Table 3: Evaluation of over-translation. The lower the score, the better the system deals with the over-translation problem.

| System | AER | SAER |
|---|---|---|
| DL4MT | 50.07 | 63.42 |
| VNMT | 49.23 | 62.28 |
| VRNMT(-TD) | 49.95 | 63.17 |
| VRNMT | **48.11** | **61.24** |

Table 4: Evaluation of word alignment quality. The lower the score is, the better word alignments are.

**Results on Source Sentences with Different Lengths**

Further, we carried out experiments to investigate our model on different groups of the test sets, which are divided according to the lengths of source sentences. Figure 3 shows that our system outperforms the others over sentences with different length spans.

**Analysis on Over Translation**

As mentioned in (Tu et al. 2016), over-translation is one of big challenges for NMT. Here we followed Zhang et al. (2017) to evaluate over-translations generated by different NMT models. Concretely, we directly used *N-Gram Repetition Rate* (N-GRR) metric (Zhang, Xiong, and Su 2017) to calculate the portion of repeated n-grams in a sentence as follows:

$$\text{N-GRR} = \frac{1}{C \cdot R} \sum_{c=1}^{C} \sum_{r=1}^{R} \frac{|\text{N-grams}_{c,r}| - |u(\text{N-grams}_{c,r})|}{|\text{N-grams}_{c,r}|} \tag{27}$$

where $|\text{N-grams}_{c,r}|$ is the number of total n-grams in the $r$-th translation of the $c$-th sentence in the testing corpus, and $|u(\text{N-grams}_{c,r})|$ denotes the number of n-grams after duplicate ngrams are removed. By comparing N-GRR scores of translations against those of references, we can roughly know how serious the over-translation problem is. Table 3 gives the final results. We find that our model is able to better deal with over-translation issue than other models.

**Analysis on Attention Results**

The attention model heavily depends on target-side hidden state vectors, which are in turn dependent on the previous latent random variables in our model, as illustrated in Eq. (22)-(25). Therefore, if latent variables are helpful for the calculation of target-side hidden state vectors, the attention model can also be improved accordingly. To testify this, we conducted experiments on the evaluation dataset provided by Liu and Sun (2015), which contains 900 manually aligned Chinese-English sentence pairs. Specifically, we first forced the decoder to output reference translations so as to obtain word alignments between input sentences and their reference translations according to attention weights. Then, we used the alignment error rate (AER) (Och and Ney 2003) and the soft version (SAER) of AER (Tu et al. 2016) to evaluate alignment performance. From Table 4, we can conclude that the incorporation of latent variables also improves the attention model as expected.

**Case Study**

To understand why our model outperforms the others, we compared and analyzed their 1-best translations. Table 2 provides a translation example with its various translations.

| System | BLEU |
|---|---|
| BPEChar | 23.9 |
| RecAtten | 25.0 |
| ConvEncoder | 24.2 |
| Moses | 20.54 |
| DL4MT | 24.88 |
| VNMT | 25.49 |
| VRNMT(-TD) | 25.34 |
| VRNMT | **25.93**$^{*}_{++}$ |

Table 5: Case-sensitive BLEU scores of English-German translation. We directly displayed the results of the first three models provided in (Gehring et al. 2017). **BPEChar** (Chung, Cho, and Bengio 2016) presented a character-level decoder for NMT, **RecAtten** (Yang et al. 2017) introduced a recurrent attention model to better capture source-side context for NMT, and **ConvEncoder** (Gehring et al. 2017) explored the convolutional encoder to encode the source sentence.

We have found that the translation produced by Moses is non-fluent than those of NMT systems. In addition to the issues of incorrect translation and over-translation, the first three NMT systems (DL4MT, VNMT, VRNMT(-TD)) do not adequately convey the meaning of the source sentence to the target as some source phrases have not been translated at all, such as "*wèi cǐ yī wēijī (for this crisis)*". By contrast, due to the advantage of modeling long-distance dependencies among target words, VRNMT is able to produce a more complete, fluent, and accurate translation.

### 4.4 Results on English-German Translation

We also carried out experiments on English-German translation. Results are shown in Table 5. We provided results of previous work (Chung, Cho, and Bengio 2016; Yang et al. 2017; Gehring et al. 2017) on this dataset too.

Specifically, VRNMT still outperforms Moses, DL4MT, VNMT, VRNMT(-TD), achieving gains of 5.39, 1.05, 0.44 and 0.59 BLEU points. Additionally, VRNMT reaches the performance level that is competitive to or higher than several recent NMT systems. Note that our approach is orthogonal to these previous models. Therefore it can be adapted to these models. We leave this adaptation to our future work.

## 5. Related Work

The previous studies that are related to our work mainly include NMT and variational neural models.

**NMT**. Most NMT models focus on how to translate a source sentence to a target sentence with an encoder-decoder neural network (Kalchbrenner and Blunsom 2013; Cho et al. 2014; Sutskever, Vinyals, and Le 2014). To handle the defeat of encoding all source-side information into a fixed-length vector, Bahdanau et al. (2015) proposed attention-based NMT, which has now become the dominant architecture. However, this model usually suffer from attention failures, which usually lead to undesirable translations. Therefore, many researchers then resorted to better attention mechanisms (Luong, Pham, and Manning 2015; Cheng et al. 2016; Tu et al. 2016; Feng et al. 2016; Meng et al. 2016; Calixto, Liu, and Campbell 2017), or more effective neural networks (Wang et al. 2016; Gehring et al. 2017; Wang et al. 2017), or exploiting external knowledge (Chen et al. 2017; Li et al. 2017; Zhang et al. 2017). All these models are designed within the discriminative encoder-decoder framework, leaving the explicit exploration of underlying semantics an open problem. To combine the strengths of discriminative and generative modeling, Zhang et al. (2016) presented VNMT that incorporates a continuous latent variable to model the underlying semantics of sentence pairs.

**Variational Neural Networks**. Kingma et al. (2014) as well as Rezende et al. (2014) focused on variational neural networks, which are effective in the inference and learning of directed probabilistic models on large-scale dataset. Typically, these models introduce a neural inference model to approximate the intractable posterior, and optimize model parameters jointly with a reparameterized variational lower bound. Further, Kingma et al. (2014b) adapted these models to semi-supervised learning. Chung et al. (2015) incorporated latent variables into the hidden states of a recurrent neural network, while Gregor et al. (2015) combined a novel spatial attention mechanism that mimics the foveation of human eyes, with a sequential variational auto-encoding framework that allows the iterative construction of complex images. Miao et al. (2016) proposed a generic variational inference framework for generative and conditional models of text.

Both (Zhang et al. 2016) and (Chung et al. 2015) are the most related to our work. In our model, we extended VNMT (Zhang et al. 2016) to a recurrent framework, which has been proven to be more effective for machine translation. Besides, different from (Chung et al. 2015) that work on speech generation and handwriting generation, we introduces a sequence of recurrent latent variables for the semantic modeling of NMT, which, to the best of our knowledge, has never been investigated before.

## 6. Conclusions and Future Work

This paper has presented a variational recurrent NMT model that introduces a sequence of continuous latent variables to capture the underlying semantics of sentence pairs. Similar to VNMT, we approximate the posterior distribution with neural networks and reparameterize the variational lower bound. In doing so, our model becomes an end-to-end neural network which can be optimized through the stochastic gradient algorithms. Compared with the dominant NMT and VNMT, our model not only captures the global semantic contexts but also models strong and complex dependencies among generated words at different timesteps. Experiments on Chinese-English and English-German translation tasks demonstrate the effectiveness of our model.

Our future works include the following aspects. We will study how to better exploit latent variables to further improve NMT. Additionally, we are also interested in applying our model to other similar tasks using encoder-decoder framework, such as neural text summarization, neural dialogue generation.

## Acknowledgments

The authors were supported by National Natural Science Foundation of China (Nos. 61672440, 61622209 and 61573294), Scientific Research Project of National Language Committee of China (Grant No. YB135-49). We also thank the reviewers for their insightful comments.